\documentclass[twoside]{article}

\usepackage{aistats2021}
\usepackage{fancyhdr}
\usepackage[algoruled,vlined]{algorithm2e}
\usepackage{amsmath} 
\usepackage{bm}      
\usepackage{graphicx} 
\usepackage{comment}
\usepackage{epstopdf}
\usepackage{cite}
\usepackage{amsmath,amsthm}
\usepackage{amssymb}
\usepackage{enumerate}

\usepackage{multirow}

\DeclareMathOperator*{\argmin}{argmin}

\newcommand{\D}{\mathbf{D}}
\newcommand{\Q}{\mathbf{Q}}
\renewcommand{\d}{\mathbf{d}}

\newcommand{\I}{\mathbf{I}}

\newcommand{\U}{\mathbf{U}}
\newcommand{\V}{\mathbf{V}}

\renewcommand{\P}{\mathbf{P}}

\renewcommand{\H}{\mathbf{H}}

\newcommand{\x}{\mathbf{x}}
\newcommand{\y}{\mathbf{y}}
\newcommand{\z}{\mathbf{z}}
\newcommand{\X}{\mathbf{X}}

\newcommand{\Y}{\mathbf{Y}}
\newcommand{\Z}{\mathbf{Z}}

\newcommand{\phib}{\boldsymbol{\phi}}
\newcommand{\Phib}{\boldsymbol{\Phi}}

\renewcommand{\raggedright}{\leftskip=0pt \rightskip=0pt plus 0cm}
\newcommand{\tabincell}[2]{\begin{tabular}{@{}#1@{}}#2\end{tabular}}

\newcommand{\arcarrow}[8]
{   \pgfmathsetmacro{\rin}{#1}
	\pgfmathsetmacro{\rmid}{#2}
	\pgfmathsetmacro{\rout}{#3}
	\pgfmathsetmacro{\astart}{#4}
	\pgfmathsetmacro{\aend}{#5}
	\pgfmathsetmacro{\atip}{#6}
	\fill[#7] (\astart:\rin) arc (\astart:\aend:\rin) -- (\aend+\atip:\rmid) -- 
	(\aend:\rout) arc (\aend:\astart:\rout) -- (\astart+\atip:\rmid) -- cycle;
	\path[decoration={text along path, text={#8}, text align={align=center}, 
		raise=-0.5ex},decorate] (\astart+\atip:\rmid) arc (\astart+\atip:\aend+\atip:\rmid);
}
\title{\textbf{SparGE: Sparse Coding-based Patient Similarity Learning via Low-rank Constraints and Graph Embedding}}
\author{
 Xian Wei $^\ast $, See Kiong Ng $^1$, Tongtong Zhang $^2$, Yingjie Liu $^3$}
\date{{\tt\small xian.wei@tum.de $^\ast $}, {\tt\small seekiong@nus.edu.sg $^1$}, {\tt\small tongtong$\_$zhang@sjtu.edu.cn $^2$}, {\tt\small yjLiu6@hotmail.com$^3$}
}
\begin{document}

    
\maketitle

\begin{abstract}
Patient similarity assessment (PSA) is pivotal to evidence-based and personalized medicine, enabled by analyzing the increasingly available electronic health records (EHRs).  However, machine learning approaches for PSA  has to deal with inherent data deficiencies of EHRs, namely missing values, noise, and small sample sizes.
	In this work, an end-to-end discriminative learning framework, called \emph{SparGE}, is proposed to address these data challenges of EHR for PSA.
	\emph{SparGE} measures similarity by jointly sparse coding and graph embedding.
	First, we use low-rank constrained sparse coding to identify and calculate weight for  similar patients, while denoising against missing values.
	Then, graph embedding on sparse representations is adopted to measure the similarity between patient pairs via preserving local relationships defined by distances.
	Finally, a global cost function is constructed to optimize related parameters. 
    Experimental results on two private and public real-world healthcare datasets, namely \textit{SingHEART} and MIMIC-III,
     show that the proposed \emph{SparGE} significantly outperforms other machine learning patient similarity methods. 		
\end{abstract}

\section{INTRODUCTION}
\label{sec:01} 

Evidence-based and personalized medicine is crucial for healthcare transformation in quality, for which patient similarity assessment (PSA) plays a prominent part.  Measuring how similar a pair of patients are according to their health records, PSA helps to improve the quality of clinical decision making without incurring additional efforts from doctors.  In fact, PSA is practised by experienced human doctors who often make their clinical decisions by incorporating their medical knowledge,  with their prior experience on patients who are diagnosed with similar diseases or showing similar symptoms.

The increasing availability of electronic health records (EHRs) which are now collected as part of routine care provides an unprecedented opportunity for patient similarity assessment through machine learning, which can further enable the automation of personalized medicine \cite{zhang2014person, Lee2015Personalized_mortality_PLoS}, 
disease classification \cite{sharafoddini2017patient_jmir}, trajectory prediction \cite{Ebadollahi2010}, 
cohort study \cite{che2017rnn_sdm}, medical diagnosis \cite{gottlieb2013} and more.  Patient similarity learning involves learning an effective mathematical representation of the patients based on the information in the EHRs, and  learning a reliable similarity metric between the patients.

However, as EHRs consist of heterogeneous and high-dimensional multi-source data elements, including diagnosis codes, lab results, prescription data, patient-reported symptoms, and so on, there are inherent data challenges of EHRs  that patient similarity learning needs to address,  
 specifically:  \romannumeral 1) Missing values: EHR data typically consist of a high rate of missing values, as patients have different sets of lab tests and medical conditions;  
\romannumeral 2) Noise: EHRs are also innately noisy, especially when patient-reported data; and \romannumeral 
3) Sparsity: EHR consist of data elements that are high-dimensional which require large  patient sample sizes especially with data-hungry machine learning methods.

  


  
To tackle these challenges, we  
construct the patient similarity metric by appealing to the principle of sparsity, using 
an end-to-end discriminative learning framework that jointly learns low rank constrained \emph{Spar}se coding and \emph{G}raph \emph{E}mbedding
(\emph{SparGE}) for patient similarity. 
\emph{SparGE} learns a similarity measurement by sequentially performing low-rank constrained sparse coding and graph embedding: 
low-rank constrained sparse coding is used for finding and weighting similar patients, as well as denoising against missing values, while
graph embedding on sparse representation is used for constructing distances to  measure the similarity between patient pairs.

\subsection{Related Works}
Various approaches have been proposed for patient similarity. 
The classical approaches used standard metrics such as Euclidean distance in raw data space
to measure the similarity among patients \cite{sharafoddini2017patient_jmir, jia2020patient_jmi}, but the raw data often have rich redundant and heterogeneous information with unknown distribution. 

%
As such, most works transformed the raw data into a novel representation space, 
where various types of similarity metrics are constructed to describe the similarity between patients.
The typical transformations include kernel function \cite{chan2010machine_BIBMW},  and 
dimensionality reduction methods, such as Principal Component Analysis (PCA), and factor analysis \cite{parimbelli2018patient_jbi}.
For example, \cite{wang2011semisupervised_TC} learned a distance metric in  low-dimensional embedding of data where the data points that are to be linked are pushed as close as possible, while the points that are not to be linked are pulled away as far as possible. 
Similar approaches were also proposed in 
\cite{sun2010localized_ICPR, sun2012supervised_acm, wang2011imet_icdm, wang2012medical_icpr, wang2015psf_JBHI}.
These approaches relied on the construction of relation matrix in raw data space,  and then convey it from high-dimensional raw data space to lower dimensions. 
Finding an optimal distance metric is then equivalent to finding an optimal dimension reduction projection matrix.
However, these approaches often suffer from  low efficiency in extracting nonlinear discrimination among data due to their shallow linear projection architecture.

Recent developments in deep learning have shown that constructing metrics in deep representation space can improve discriminative patient similarity learning 
\cite{miotto2016deep_patient_SR, pai2018patient_jmb, zhu2016measuring_concept_icdm, pai2019netdx_msb}.  However,
such approaches often require not only special computing hardware which may not (yet) be available in a traditional hospital setting, but also a very 
large amount of training data and prohibitive training efforts as compared to shallow learning approaches.
The aforementioned methods are also vulnerable to noise and missing values, which are prevalent in EHR data.  

Some researchers have  looked into measuring patient similarity based on the concept of sparsity. For example, the 
methods in \cite{suo2018multi_icdm, zhan2016low_icdm, zhou2012modeling_sigkdd} imposed the sparse regularization on dimension reduction projection to 
reduce  irrelevant and redundant information, but they did not  consider the robustness to noise corruption. 
%
Also, most of the methods mentioned above are not competent in handling incomplete input matrices\cite{review_EHR_all}. Most research \cite{sun2010system_missing1_icdm, chattopadhyay2008suicidal_missing2_icadma} 
 opt to fill the data matrix with  imputation approaches. On the other hand,
low-rank matrix completion methods have been developed for solving missing value problem \cite{missing1,missing2}, which inspire us to utilize low-rank tools for this work.

Another common inadequacy of the current models is that they neglect the underlying relationships between the data elements during the learning process. Many of the coded values in EHRs using standards such as the ICD-9 code\cite{review_EHR_all} have underlying relationships but these useful relationships are not captured and exploited by most of metric learning methods.  We propose to capture the underlying structural relationships between the various data value elements using graph embedding to exploit the structural relationship to address the innate data challenges for patient similarity learning.  In short, this paper proposes a similarity learning scheme SparGE using low-rank sparse coding to handle the problem of noise and missing values, while preserving the innate relational structure of the data elements by graph-embedding constraint.

\section{FINDING SIMILAR PATIENTS VIA LOW RANK CONSTRAINED SPARSE CODING}
\label{sec:02}

%
As mentioned previously, patient similarity learning involves two aspects: 
\romannumeral 1) finding and weighting similar patients, which is discussed in this section, and
\romannumeral 2) explicitly measuring the distance between similar patients, which will be discussed in Section~\ref{sec:03} and Section~\ref{sec:04}. 

In this work, we find similar patients by applying sparse coding with low-rank approximation of the dictionary.
The low-rank approximation of the dictionary prevents the outliers and missing values from degrading the model, while the sparse representation is for selecting similar patients while being robust to random Gaussian noise.

Let the vector $\x_i = [\x_i^{1}, \x_i^{2}, \cdots, \x_i^{m}]\in \mathbb{R}^{m}$ contain all measurable EHR variables
for the $i^{\text{th}}$ patient, with $m$ being the number of all biomarkers.
 Let $\X = [\x_1, \x_2, \cdots, \x_n]\in \mathbb{R}^{m \times n}$
be a collection of $n$  patients' raw EHR data points, potentially with excessive noises.

The low-rank approximation approach aims to find a low-rank matrix $\Z$ that is as close to $\X$ as possible 
\cite{candes2011robust_jacm, li2019survey_arxiv},
by optimizing the following minimization problem

\begin{equation}
	\label{eq_LR}
	\mathop{\arg\min}_{\Z} \|\Z_{\Omega} - \X_{\Omega} \|_F^2 + \lambda \|\Z\|_{*}, %
\end{equation}
where $\|\Z\|_{*}$ denotes the nuclear norm of the matrix $\Z$, 
$\lambda \in \mathbb{R}^{+}$ weights the low-rank regularization against the reconstruction error, 
$$\Z_{\Omega} = \H_{\Omega} \odot \X + \mathbf{N}.$$
Therein, $\odot$ is the element-wise multiplication operator, 
${\Omega}$ indicates the support of observed entries,
$\H_{\Omega}$ and $\mathbf{N}$ are the sampling matrix and noise matrix, respectively.

Various works have observed that 
 data can have strong \textit{self-expressiveness} property 
\cite{wright2009robust_pami, gao2012dimensionality_prl, elhamifar2013sparse_pami, wei_trace_2020_pami},
i.e., each data point can be efficiently represented as a linear or affine combination of some selected points in the dataset. 
The underlying assumption of \textit{self-expressiveness} property is that an incoming new
data point can be embedded into a subspace spanned by a selected subset of the dataset.  In this work, we follow the observation and assume that 
the EHR data of a patient can be expressed by other patients, which encapsulates the fundamental assumption of patient similarity assessments. 

Let $\Z$, i.e., the low-rank approximated matrix of $\X$, 
be a self-expressive \textit{dictionary} for all patients except the test patient,
in which each column vector (or point) contains the EHR data for a patient, and that
each column vector can be written as a linear combination of the other column vectors.
%
Given a new unlabelled test patient $\z$, we wish to represent it in a subspace spanned by 
a subset from $\Z$, i.e., finding only a few elements in $\Z$ to approximate $\z$.
Formally, this can be described as
\begin{equation}
\label{eq_SR0}
\z = \Z\bm{\phi}, %
\end{equation}
where $\bm{\phi} \in \mathbb{R}^{n}$ is the sparse representation of $\z$, i.e.,most entries are zero in $\bm{\phi}$.
Hereby, $\bm{\phi}$ corresponds to adaptively select and weight a few elements from $\Z$.
Each absolute value in $\bm{\phi}$  reflects the importance of the selected element (patient) for approximating the 
test patient $\z$, i.e., the similarity between the selected patient and the test patient. 

With a fixed dictionary $\Z$, there are several ways of finding the sparse representation $\bm{\phi}$. 
One popular solution of the sparse learning problem above 
is given by solving the following minimization problem
\begin{equation}
\label{eq_sr}
\phib^{*} := \operatorname*{argmin}_{\phib \in \mathbb{R}^{n}}
\tfrac{1}{2} \| \z - \Z \phib \|_{2}^{2} + g(\phib).
\end{equation}
%
%
Therein, the first term penalizes the reconstruction error of sparse representations, and
the second term is a sparsity promoting regularizer.
Often, the function $g$ is chosen to be separable, i.e., its evaluation is computed 
as the sum of functions of the individual components of its argument.
%
There are many choices for $g$ in the literature, such as the $\ell_{0}$-(quasi-)norm, $\ell_{1}$-(quasi-)norm  
and their variations \cite{bach2012structured_SS, Wei2017Book}, which have promoted the global sparse structures 
of $\phib$.

Finally, given input data $\X$, by combining the Eq.~\eqref{eq_LR} and Eq.~\eqref{eq_sr}, 
the low-rank approximation based sparse coding problem can be formulated as 
\begin{equation}
	\label{eq_SR_LR}
    \operatorname*{min}_{\phib \in \mathbb{R}^{n}, \Z}
	\tfrac{1}{2} \| \z - \Z \phib \|_{2}^{2} + g(\phib) + \lambda_1 \|\Z_{\Omega} - \X_{\Omega}\|_F^2 + \lambda_2 \|\Z\|_{*}.
\end{equation}

The two goals of ~\eqref{eq_SR_LR} are to:
\romannumeral 1) Reconstruct the clean data  $\Z$ from raw input $\X$;
\romannumeral 2) Given a query patient $\z$, find minimal number of patients (the sparsity of $\phib$), to express the query patient.
The  non-zero entries in $\phib$ denote the indices of similar patients
and their values reveal the linear dependence between the query patient and selected similar patients.

\section{JOINT LEARNING OF LOW-RANK CONSTRAINED DICTIONARY AND GRAPH EMBEDDING}
\label{sec:03}
In Section \ref{sec:02}, we treat data as the predefined dictionary, and the success of Eq.~\eqref{eq_SR0} and Eq.~\eqref{eq_sr} 
is thus heavily dependent on the \textit{self-expressiveness} property of the data.
However, there is no guarantee that predefined basis vectors (dictionary) will be well-matched to the structure of test patient data.
In Section in Section \ref{sec:041}, we propose that a low-rank constrained dictionary be learned from the data so as to have more flexibility to adapt the sparse representation to the patients' EHR data. Then,  in Section \ref{sec:042},  we apply  dimension reduction  on the sparse representations to calculate the distance of patient pairs  in low-dimensional space, in order to explicitly measure the similarity between patients.
Finally, in Section \ref{sec:04}, a global cost function is constructed to jointly learn a low-rank constrained dictionary and a projection for graph embedding.

\subsection{Low-rank Constrained Dictionary Learning}
\label{sec:041}
%
Formally, let $\D:= [\d_{1}, \ldots, \d_{k}]\in \mathbb{R}^{m\times k}$ be a shared dictionary 
that adaptively represents a signal as a sparse vector ${\phib} \in \mathbb{R}^{k}$, i.e.,
\begin{equation}
	\label{eq_DL_framework}
	\z = \D {\phib}.
\end{equation}
Dictionary learning aims to find a collection of \emph{atoms} $\d_{i} \in
\mathbb{R}^{m}$ for $i = 1, \ldots, k$, such that each data point can be
approximated by a linear combination of a small subset of atoms in $\D$. 
Given a collection of $n$ data points in $\mathbb{R}^{m}$ as $\Z := [\z_{1}, \ldots, \z_{n}] \in \mathbb{R}^{m \times n}$,
\begin{equation}
	\label{eq_DL_framework2}
	\Z = \D \Phib,
\end{equation}
with $\Phib := [\phib_1, \ldots, \phib_n] \in \mathbb{R}^{k \times n}$. 
Based on Eq.~\eqref{eq_DL_framework2}, 
$$\text{rank}(\Z) = \min \{\text{rank}(\D), \text{rank}(\Phib)\}.$$
In other words, the low rank of $\D$ results in the low rank of $\Z$.
The solution of the above dictionary learning problem, associated with the low-rank approximation, 
is given by solving the following minimization problem:
\begin{equation}
\label{eq_sparse_coding}
\operatorname*{min}_{\Z, \D, \Phib}
\tfrac{1}{2} \| \Z - \D \Phib \|_{F}^{2} + \sum_{i = 1}^{n} g(\phib_{i}) + \lambda_1 \|\Z_{\Omega} - \X_{\Omega}\|_F^2  + \lambda_2 \|\D\|_{*},
\end{equation}
Therein, the weighting factors $\lambda_1, \lambda_2\in \mathbb{R}^{+}$
control the influence of two parts on the final solution,
$\D \in \mathcal{D}$ with
\begin{align*}
	\label{constraints_dictionary_CDL}
	\mathcal{D}:=	\left\{ \mathbf{D} \in \mathbb{R}^{m \times k} \big| \|\mathbf{d}_{i}\|_{2} \leq 1 \right\}.
\end{align*}

Once $\D$ is given,  Problem~\eqref{eq_sparse_coding} can be rewritten as  the following 
decoupled sample-wise \emph{sparse regression} problem.
Specifically, for each sample $\z$, we have:
\begin{equation*}
\label{eq_group_sparse_regression}
\min_{\phib \in \mathbb{R}^{k}, \Z} 
f_{\x}(\phib, \Z) := \tfrac{1}{2} \| \z - \D \phib \|_{2}^{2} + g(\phib) + \lambda_1 \|\Z_{\Omega} - \X_{\Omega}\|_F^2 .
\end{equation*}

The solution of the sample-wise sparse regression problem can be treated as a function in $\z$, i.e., 
\begin{equation}
\label{eq_sparse_regression_solution}
\phib_{\D}(\x, \z) := \operatorname*{argmin}_{\phib \in \mathbb{R}^{k}} f_{\x}(\phib, \Z)
\end{equation}
The implicit function $\phib_{\D}(\x, \z)$ is locally differentiable with respect to $\D$, $\x$ and $\z$.
 We refer the reader to the Supplementary Materials for more details.
Such a property allows us to integrate $\phib_{\D}(\x, \z)$ into a larger learning architecture. 

Furthermore, as suggested by the works of
\cite{mair:pami12, elhamifar2013sparse_pami, wei2016trace_cvpr},
the sparse representation above can be further processed to reveal the underlying discriminative information 
which benefits both supervised and unsupervised learning tasks. 
In other words, once the sparse representations are calculated for all patients' EHR data via Eq.~\eqref{eq_sparse_regression_solution}, we can assume that 
similar patients  have similar sparse structures. 
However, the explicit measurement of the pairwise similarity is not clearly constructed.
In the next section, we show how to learn an explicit similarity measurement in the sparse domain.

\subsection{Learning Similarity Measurement in Sparse Domain}
\label{sec:042}
%
With the sparse representations of all samples at hand, specific similarity learning 
algorithms can be directly applied to these sparse coefficients to 
extract further discriminative information between patients.
We can learn a patient similarity measurement in the sparse domain as follows.

%
The \textit{generalized Mahalanobis distance} is often used 
for the patient similarity problem \cite{suo2018multi_icdm, wang2011imet_icdm, wang2015psf_JBHI}. 
In this work, we construct a \textit{generalized Mahalanobis distance} in the sparse domain 
\begin{equation}
\label{eq_similarity}
d^2(\phib_i, \phib_j) = (\phib_i - \phib_j)^\top \P (\phib_i - \phib_j)
\end{equation}
%
to measure the similarity between $\phib_i$ and $\phib_j$.
Therein, ${\P} \in \mathbb{R}^{k\times k}$ is a \textit{Symmetric Positive Semi-Definite} (SPSD) matrix to be learned.

In consideration of the difficulty of directly learning ${\P}$, it is reasonable to assume that the similarity is measured 
in an embedded space of sparse representations via 
\begin{equation}
\label{eq_DR} 
\y_i = \mathbf{U} ^\top \phib_i, 
\end{equation}
%
where $\mathbf{U} \in \mathbb{R}^{k\times l}$ is an dimension reduction transformation with $l \ll k$, 
$\y_{i} \in \mathbb{R}^{l}$ is a representation in the low-dimensional embedded space.
Since ${\P}$ is a SPSD matrix, $\mathbf{U}$ could be viewed as a component for approximating ${\P}$ with
$$ {\P} = \mathbf{U}\mathbf{U}^\top.$$
%
%
Then, the Eq.~\eqref{eq_similarity} could be rewritten as 
\begin{equation}
\begin{split}
\label{eq_similarity_re}
d^2(\phib_i, \phib_j) & =  (\mathbf{U}^{\top} \phib_i - \mathbf{U}^{\top} \phib_j)^{\top} (\mathbf{U}^{\top} \phib_i - \mathbf{U}^{\top} \phib_j) \\
&  = (\y_i - \y_j)^\top  (\y_i - \y_j) \\
&  = \|\y_i - \y_j\|_2^2 \\
&  = d^2(\y_i, \y_j) 
\end{split}
\end{equation}
The \textit{generalized Mahalanobis distance} in sparse domain could be obtained
by computing the Euclidean distance in $\mathbb{R}^{l}$.
Once $\{\phib_i \}$ and $\U$ are given, $\{\y_i\}$ are computed by Eq.~\eqref{eq_DR}.
Hence, finding $k$ similar patients is achieved by applying Nearest Neighbourhood method on $\{\y_i\}$.
Note that from Eq.~\eqref{eq_similarity_re},  the key is to learn the parameters $\Z$, $\D$ and $\U$.
%
We will propose a unified cost function, for jointly learning  $\Z$, $\D$ and $\U$,
with both the supervised and the unsupervised settings in the next section.

\subsection{The Proposed \emph{SparGE} Method}
\label{sec:04}
The previous sections introduce the low-rank constrained dictionary and the patient similarity measurement.
In this section, 
we design a cost function that jointly learns 
both the low-rank constrained dictionary and an orthogonal transformation, 
to efficiently extract low-dimensional features on sparse representations. 

\subsubsection{A Generic Joint Learning Framework}

Let us denote by $\mathbf{X} := [\mathbf{x}_1, \ldots, \mathbf{x}_n] \in \mathbb{R}^{m\times n}$ the given data matrix  
containing $n$ high-dimensional data samples, and $\mathbf{Y} := [\mathbf{y}_1, \ldots, \mathbf{y}_n] \in 
\mathbb{R}^{{l}\times n}$ with ${l}<m$ being its corresponding unknown low-dimensional representation 
via some mapping 
\begin{equation}
	\label{eq_mapping}
	\mathcal{L} \colon \mathbf{x}_{i} \mapsto \mathbf{z}_{i} \mapsto \phib_{i} \mapsto \y_{i}, \quad\qquad\text{for all~}i=1,\ldots,n, 
\end{equation}
where $\mathbf{x}_{i}, \mathbf{z}_{i}, \phib_{i}, \y_{i}$ denote the input signal, the recovered clean signal, 
the sparse vector, and the low-dimensional representation, respectively.


To achieve computational efficiency,
many classic linear dimension reduction methods restrict the row vectors in $\mathbf{U}$  to be orthonormal,
{i.e.}, $\mathbf{U}$  belongs to the Stiefel manifold
\begin{equation}
\label{eq_St}
\mathcal{U} := \{\mathbf{U}\in \mathbb{R}^{k\times l}| \mathbf{U}^\top \mathbf{U} = \mathbf{I}_l \}.
\end{equation}

The loss introduced by the mapping $\mathcal{L} \colon \mathbf{x}_{i} \mapsto \mathbf{z}_{i} \mapsto \phib_{i} \mapsto \y_{i}, \forall i$ 
can then be measured by the function
\begin{equation}
	\begin{split}
		\label{eq_trace_quotient_dictLearn}
		& \operatorname*{argmin}_{\Z, \D, \U} \; \mathcal{L}_{(\mathbf{Z}, \mathbf{D}, \mathbf{U})} (\mathbf{X};\mathbf{Y}) \\
		& \text{subject to}, \; \phib_{\D}(\x, \z) := \operatorname*{argmin}_{\phib \in \mathbb{R}^{k}} f_{\x}(\phib, \Z)
	\end{split}
\end{equation}
which jointly learns the clean data, the dictionary and the linear orthogonal projection,  combining the Graph Embedding formulation in Eq.~\eqref{eq_trace_quotient_st}
and the sparse coding formulation in Eq.~\eqref{eq_sparse_coding} to be described in the next section.
Graph Embedding is well known as an efficient tool to extract local similarity of data points.
In this way, \emph{SparGE}
jointly learns low-rank constrained sparse coding and graph embedding for patient similarity.
%
%

%
%
 In the next section, we give two examples to evaluate the problem\eqref{eq_trace_quotient_dictLearn}, but it is important to note that the proposed \emph{SparGE}  is a flexible and generic joint learning framework and thus  not limited to two methods described. 

 Considering that $ \mathcal{L}$ in~\eqref{eq_trace_quotient_dictLearn} is a smooth function and has the first-order derivatives to $\D, \U$,
 the simple gradient method can be used for optimizing the function.
 In the work, we apply the alternating optimization framework \cite{boyd2010distributed_ML} to solve the proposed minimization problem \eqref{eq_trace_quotient_dictLearn},
 i.e.,  solve for one of parameters if the others are considered fixed.
 A generic gradient algorithm is summarized in Algorithm~\ref{algo_SPARGE}.
We refer the reader to the Supplementary Materials for more details.

The formulations in Eq.~\eqref{eq_LR}, Eq.~\eqref{eq_SR_LR} and Eq.~\eqref{eq_sparse_coding} 
involve minimizing the nuclear norm $\min \|\Q\|_{*}$, i.e.,
 computing the low-rank approximation $\Q^{\star}$ to a given matrix $\Q$.
It can be decomposed using the famous \textit{nuclear norm minimization} \cite{candes2011robust_jacm, li2019survey_arxiv},
which repeatedly shrinks the singular values of an appropriate matrix.
One simple method is Singular Value Thresholding (SVT) \cite{cai2010singular_siam, recht2010guaranteed_siam}. 
Formally, let $\Q$ admit an economic singular value decomposition (SVD) as $\Q = \U\Sigma \V^\top$
where $\U$ are the left singular vectors, $\V$ are the right singular vectors, 
and $\Sigma$ is a diagonal matrix and has singular values on the diagonal. 
Performing SVT on $\Q$ is denoted by
$$SVT(\Q) = \U S_{\zeta}(\Sigma) \V^\top,$$ 
where 
\begin{align}
	\label{eq_SVT}
	S_{\zeta}(x) :=
	\left\{ \begin{array}{ll} 
		x-\zeta,   \mbox{ if $x>\zeta$}\\
		0,  \mbox{otherwise}\\
	\end{array}\right.
\end{align}
%
with $\zeta$ being a constant as threshold for singular values.
\begin{algorithm}[h]
	\caption{A \emph{SparGE} Algorithm.} 
	\label{algo_SPARGE} 
	\SetAlgoNoLine
	\SetKwHangingKw{PRE}{Preprocessing~:}
	\SetKwHangingKw{IN}{Input~:}
	\SetKwHangingKw{Sa}{Step~1:}
	\SetKwHangingKw{Sb}{Step~2:}
	\SetKwHangingKw{Sc}{Step~3:}
	\SetKwHangingKw{Sd}{Step~4:}
	\SetKwHangingKw{Se}{Step~5:}
	\SetKwHangingKw{Sf}{Step~6:}
	\SetKwHangingKw{OUT}{\hspace{-1.6mm}Output:}

	\IN{Given training set $\textbf{X} \in \mathbb{R}^{m \times n}$, 
		parameters $\lambda_1$, $\lambda_2$, $\gamma$, $\zeta$ }
	
	\OUT{$\mathbf{Z}^{*},\mathbf{D}^{*},\bm{\Phi}^{*}, {\U}^{*}$}		
	
	\Sa{Initialize $\mathbf{Z}^{(0)}$ by applying SVT on $\X$. Then, learn $\mathbf{D}^{(0)}$ from $\mathbf{Z}^{(0)}$.
		Finally, compute $\bm{\Phi}^{(0)}$, ${{\U}}^{(0)}$, given fixed $\mathbf{Z}^{(0)}$ and $\mathbf{D}^{(0)}$
	\vspace{1mm}}

	\Sb{Set $j=j+1$ \vspace{1mm}, fix $\mathbf{Z}^{(j)}$, $\mathbf{D}^{(j)}$,
		compute sparse coefficients ${\bm{\Phi}}^{(j+1)}$ 
		via Eq.~\eqref{eq_sparse_regression_solution} \vspace{1mm}}	
		
	\Sc{Compute the search direction $\mathbf{H}^{(j)} = \text{grad}_{\mathcal{L}}(\mathbf{D}^{(j)}, {{\U}}^{(j)}) $ \vspace{1mm}}
	
	\Sd{Update $( {\hat{\D}}^{(j+1)}, {\hat{\U}}^{(j+1)} ) \!\gets\!
		({{\D}}^{(j)}, {{\U}}^{(j)})  \!+\! 
		\gamma \mathbf{H}^{(j)}$. $\hat{{\D}}^{(j+1)} = SVT( \hat{{\D}}^{(j+1)} )$.
		Project ${\hat{\D}}^{(j+1)}, {\hat{\U}}^{(j+1)}$ onto ${{\D}}^{(j+1)} \in \mathcal{D}, {{\U}}^{(j+1)} \in \mathcal{U}$
		\vspace{1mm}} 
	
	\Se{Update $\mathbf{Z}^{(j+1)} = {{\D}}^{(j+1)} {\bm{\Phi}}^{(j+1)}$ \vspace{1mm}}
	
	\Sf{If $ \big\| \mathbf{H}^{(j)}  \big\|$ is small enough, stop. 
		Otherwise, go to Step 2\vspace{1mm}}
\end{algorithm}
\subsubsection{Unsupervised Graph Embedding}
\label{sec:0420}
Given a data set $\mathbf{X}_{i} = [\mathbf{x}_{i1}, \ldots, \mathbf{x}_{in_{i}}] \in 
\mathbb{R}^{m \times n_{i}}$, $i = 1, \ldots, c$, where $c > 1$ denotes the class number and 
$n_{i}$ denotes the number of data samples belonging to $i$-{th} class. 
The corresponding sparse coefficients are denoted by 
${{\bm{\Phi}}}_{i} = [\phib_{i1}, \ldots, \phib_{in_{i}}] \in \mathbb{R}^{k \times n_{i}}$, 
and ${\bm{\Phi}} = [{{\bm{\Phi}}}_{1}, \ldots, {{\bm{\Phi}}}_{c}] \in \mathbb{R}^{k \times n}$.
Let ${d}_{ij} := \exp (-\|{\phib}_i - {\phib}_j\|_2^2/t)$ denotes
the Laplacian similarity between two sparse vectors ${\phib}_i$ and ${\phib}_j$ with constant $t>0$.

For the whole dataset $\mathbf{X}$, let us define the local Laplacian matrix and the non-local Laplacian matrix 
$\mathbf{L}_{L} = {\bm{\Xi}_{L}} - \mathbf{M}_{L}$ in $\mathbb{R}^{n\times n}$,
$\mathbf{L}_{N} = {\bm{\Xi}_{N}} - \mathbf{M}_{N}$ in $\mathbb{R}^{n\times n}$, respectively.
$\mathbf{M}_{N} := \{\mathbf{d}_{ij}\}$ and $\mathbf{M}_{N} := \{0\}$ when $\bm{\phi}_i$, $\bm{\phi}_j$ are adjacent.
$\mathbf{M}_{L} := \{0\} $ and $\mathbf{M}_{L} := \{\mathbf{d}_{ij}\}$ when $\bm{\phi}_i$, $\bm{\phi}_j$ are non-adjacent.
Both $ {\bm{\Xi}_{L}}$ and ${\bm{\Xi}_{N}}$ are diagonal with 
${\bm{\Xi}_{L}}_{ii} :=  \sum_{j \neq i} {\mathbf{M}_{L}}_{ij}, \forall j$
and ${\bm{\Xi}_{N}}_{ii} :=  \sum_{j \neq i} {\mathbf{M}_{N}}_{ij}, \forall j$.
Both $\mathbf{L}_{L}$ and $\mathbf{L}_{N}$ are \textit{Symmetric Positive Semi-Definite} (SPSD).
We utilize a generic algorithmic framework with supervised form to find optimal ${\U} \in \mathcal{U}$
is formulated as a minimization problem of the so-called \emph{trace quotient} or \emph{trace ratio}, i.e.,
\begin{equation}
	\label{eq_trace_quotient_st}
	\operatorname*{argmin}_{{\U}, \Z, \D} \;
	\frac{\operatorname{tr}({\U}^{\top} {\Phib_{\X}(\Z, \D)}  \mathbf{L}_{L} {\Phib_{\X}(\Z, \D)}^\top {\U})}{
		\operatorname{tr}({\U}^{\top}{\Phib_{\X}(\Z, \D)} 
		\mathbf{L}_{N} {\Phib_{\X}(\Z, \D)}^\top {\U}) },
\end{equation}

\subsubsection{Supervised Graph Embedding}
\label{sec:0421}
It is possible to develop supervised versions of the graph embedding \eqref{eq_trace_quotient_st} by taking the class labels into account. 
Assume that there are $c$ classes of patients. 

Let $\X_{i} = [{\x}_{i1}, \ldots, {\x}_{in_{i}}] \in 
\mathbb{R}^{m \times n_{i}}$ for $i = 1, \ldots, c$ with $n_{i}$ being 
the number of samples in the $i^{\mathrm{th}}$ class. 
The corresponding sparse coefficients are denoted by 
${{\Phib}}_{i} := [{\phib}_{i1}, \ldots, {\phib}_{in_{i}}] \in \mathbb{R}^{r \times n_{i}}$, 
and ${{\Phib}} := [{{\Phib}}_{1}, \ldots, {{\Phib}}_{c}] \in \mathbb{R}^{r \times n}$
with $n = \sum\limits_{i=1}^{c} n_{i}$.  

The main idea in supervised methods is to maintain the original neighbor relations of points from the 
same class while pushing apart the neighboring points of different classes and the class labels are used to build the graph. 
%
It aims to preserve localities in such a supervised graph, will result in samples from the same class being projected close-by in the reduced space.

Let $\mathfrak{N}_{k_1}^{+}({\phib}_i)$ denote the set of $k_1$ nearest neighbors which share the same label with ${\phib}_i$, 
and $\mathfrak{N}_{k_2}^{-}({\phib}_i)$ denote the set of $k_2$ nearest neighbors among the data points whose labels 
are different to that of ${\phib}_i$.
We construct two matrices $\Z^{+} := \{z_{ij}^{+}\} \in \mathbb{R}^{n\times n}$
and $\Z^{-} := \{z_{ij}^{-}\} \in \mathbb{R}^{n\times n}$ with
\begin{equation}
	z_{ij}^{+} = \left\{\!\!
	\begin{array}{ll}
	    d(\phib_i, \phib_j), & {\phib}_j\in \mathfrak{N}_{k_1}^{+}({\phib}_i) ~\text{or}~ 
		{\phib}_i\in \mathfrak{N}_{k_1}^{+}({\phib}_j), \\
		0, & \text{otherwise},
	\end{array}
	\right.
\end{equation}
and
\begin{equation}
	z_{ij}^{-} = \left\{\!\!
	\begin{array}{ll}
		d(\phib_i, \phib_j), & {\phib}_j\in \mathfrak{N}_{k_1}^{-}({\phib}_i) ~\text{or}~ 
		{\phib}_i\in \mathfrak{N}_{k_1}^{-}({\phib}_j), \\
		0, & \text{otherwise}.
	\end{array}
	\right.
\end{equation}
Then, the Laplacian matrices for characterizing the inter-class and  
intra-class locality are defined as
\begin{equation}
	\mathbf{L}^{-} = \Y^{-} - \Z^{-}, \quad\text{and}
	\quad \mathbf{L}^{+} = \Y^{+} - \Z^{+},
\end{equation}
%
where $\Y^{+}$ and $\Y^{-}$ are two diagonal matrices defined as
%
\begin{equation}
	y^{+}_{ii} = \sum_{j \neq i} z_{ij}^{+}, \quad\text{and}\quad
	y^{-}_{ii} = \sum_{j \neq i} z_{ij}^{-}.
\end{equation}
Then, we construct the following functions
, i.e.,
\begin{equation}
	\label{eq_trace_quotient_mfa}
	\operatorname*{argmin}_{{\U}, \Z, \D} \;
	\frac{\operatorname{tr}({\U}^{\top} {\Phib_{\X}(\Z, \D)}  \mathbf{L}^{+} {\Phib_{\X}(\Z, \D)}^\top {\U})}{
		\operatorname{tr}({\U}^{\top}{\Phib_{\X}(\Z, \D)} 
		\mathbf{L}^{-} {\Phib_{\X}(\Z, \D)}^\top {\U}) },
\end{equation}

Once $\Z$, $\D$ and $\U$ are computed, the common metric, such as Euclidean distance, can be used for finding similar patients in low-dimensional embedded space, via sequentially performing Eq.~\eqref{eq_DL_framework} and Eq.~\eqref{eq_DR}. 

\section{EXPERIMENT}
\label{sec:05}
In this section, we investigate the performance of proposed methods
on two real EHR datasets.

\subsection{Experimental Settings}
For unsupervised learning experiments, we employ the K-SVD algorithm 
\cite{aharon2005k_KSVD} to compute an empirically optimal data-driven dictionary,
and initialize \emph{SparGE} algorithms. 
%
For supervised learning experiments, we adopt the same approach to generate
a sub-dictionary for each class, and then concatenate all sub-dictionaries to form 
a common dictionary.

With the given dictionary $\D^{(0)} \in \mathcal{D}$ initialized, the original  
orthogonal projection $\U^{(0)} \in \mathcal{U}$ can be directly obtained by applying 
classic TQ maximization algorithms on the sparse representations of the
samples with respect to $\D^{(0)}$. 
Apparently, when the number of training samples is huge, it is unnecessary and impractical to
perform a TQ maximization for initialization.
Thus we employ only a subset of random samples to compute the initial 
orthogonal projection $\U^{(0)}$.

More concretely, in all experiments, we set $\gamma = 10^{-3}$ and $\zeta = 0.1$ by hand 
in Eq.~\eqref{eq_trace_quotient_dictLearn} and Eq.~\eqref{eq_SVT},
the parameters for $\lambda_1, \lambda_2$ in Eq.~\eqref{eq_trace_quotient_dictLearn}
could be well tuned via performing cross validation.

Specially, for datasets without a pre-construction of training set and testing set,
all experiments are repeated ten times with different randomly constructed training set and test set, and the average of per-class recognition rates is recorded for each run. Besides, considering that each kind of data has different measurement units, all columns are presented as $m$-dimensional vectors, and normalized to have unit norm, to avoid the errors caused by different physical units.

\subsection{Singapore \textit{SingHEART} Dataset}
The Singapore \textit{SingHEART} dataset\cite{yap2019harnessing_BCD} is a private dataset that contains $683$ male and female subjects carrying 
$467$ biomarkers  from $7$ groups, with ages between  $21$–$69$ years old, between $2015$ and $2017$.

%
In our experiments, after preprocessing  data by merging synonyms and repetitive biomarkers, the total number of clinical features changes from
a total of $457$ symptoms to $383$ after data preprocessing by merging synonyms and repetitive biomarkers, and excluding subjects who lose more than $10\%$ data of their data. However, our dataset still with a lot of incomplete data, we calculate metrics for each label, and find their average weights by supporting the number of true instances for each label, the missing data are replaced by the weighted mean of their $10$ nearby measurements in each column. 
%
%
Note that the data are collected from multiple and heterogeneous sources, varying from digital numbers to texts.
%
Thereby, as part of the data preprocessing, we encode all kinds of data into digital form.
Some examples of data integer encoding results are shown in Table~\ref{table_singhealth_results}.

%

%
%

 \begin{table*}[h]
 	\begin{center}
 		\caption{ 
 			\label{table_singhealth_results} 
 			Integer Encoding examples for some multi-source \emph{SingHEART} biomarkers.
 		}
 		\begin{tabular}{c  c} 
 			\hline  
 			\tabincell{c}{ \textbf{Biomarker}}
 			&   \tabincell{c}{\textbf{Digital encoding}} 	 \\ 
 			\hline
 			\tabincell{c}{ Country of birth}	    & ``Singaporean'' $= 1$, ``Others'' $= 2$  \\
 			\hline
 			\tabincell{c}{ Marital Status}	    & ``Single'' $= 1$, ``Married'' $= 2$, ``Widowed'' $= 3$    \\
 			\hline
 			\tabincell{c}{ Religion }	    & \tabincell{c}{ ``Christian/Catholic'' $= 1$, ``Free thinker'' $= 2$,  ``Buddhist'' $= 3$, \\ ``Islam'' $= 4$, ``Hindu'' $= 5$,  ``Taoist'' $= 6$,``Others'' $= 7$}    \\
 			\hline
 			\tabincell{c}{  \tabincell{c}{ Diabetes, Hypertension, \\ Heart Attack Block, Heart Disease, \\ Hyperlipidemia, $\cdots$}}	    & ``Unknown'' $= 0$, ``Yes'' $= 1$, ``No'' $= 2$  \\
 			\hline
 			\tabincell{c}{ Smoking, Cigarettes }	    & ``Yes'' $= 1$, ``No'' $= 2$, ``Previously'' $ = 3$  \\
 			\hline
 			\tabincell{c}{ Home/Work Smoke}	    & ``Never'' $= 1$, ``Sometimes'' $= 2$, ``Most of the times'' $ = 3$, ``Never'' $ = 4$  \\
 			\hline
 			\tabincell{c}{ Alcohol, Beer, Red/White Wine}	    & ``Yes'' $= 1$, ``No'' $= 2$  \\
 			\hline
 			\tabincell{c}{ Coffee/Tea Weekly}	    & \tabincell{c}{ ``Never/rarely'' $= 1$, ``$< 1$ cup a week'' $= 2$, \\ ``$>= 1$ cup a week but $<= 1$ cup a day'' $= 3$, ``Others'' $= 4$}  \\
 			\hline
 		\end{tabular} 
 	\end{center}
 \end{table*}


\subsection{MIMIC-III Dataset}
MIMIC-III (Medical Information Mart for Intensive Care) \cite{harutyunyan2019multitask_sd} is a large public database comprised of information 
of patients in critical care units at a large tertiary care hospital. 
The dataset is used to evaluate proposed method for large-scale time series data,  
associated with $53,423$ distinct hospital admissions for adult patients (aged $16$ years or above) 
admitted to critical care units between $2001$ and $2012$. 
The task of experiments on MIMIC-III is also to predict patient mortality, according to patient similarity. 
The challenges are in two aspects: \romannumeral 1) more than $30\%$ missing values and \romannumeral 2) the temporal structure of the data.
%
In the experiments, only a subset of the MIMIC-III dataset is compiled. 
We picked about $21,142$ samples, and the first $70$ time steps of all samples that are in 48 hours. 
%
%
By using the one hot coding to the raw $17$ dimensional features, 
we get the $48 \times 76$ for each patient.
Finally, this feature matrix are reshaped as a fixed length vector to feed the machine learning algorithms.
\begin{table}[t]
\centering
	\caption{\label{table_classifiers} 
		Baseline classifiers.}
\begin{tabular}{ll}
\hline
'KNN' & $k-$Nearest Neighbor \\ \hline
'LR' & Logistic Regression  \\ \hline
'RF' & Random Forest  \\ \hline
'DT' & Decision Tree \\ \hline
'SVM' & Support Vector Machines \\ \hline
'GBDT' & Gradient Boosting Decision Tree \\ \hline
\end{tabular}
\end{table}
\subsection{Results}
\begin{figure}[t]
\centering
\includegraphics[scale=0.46]{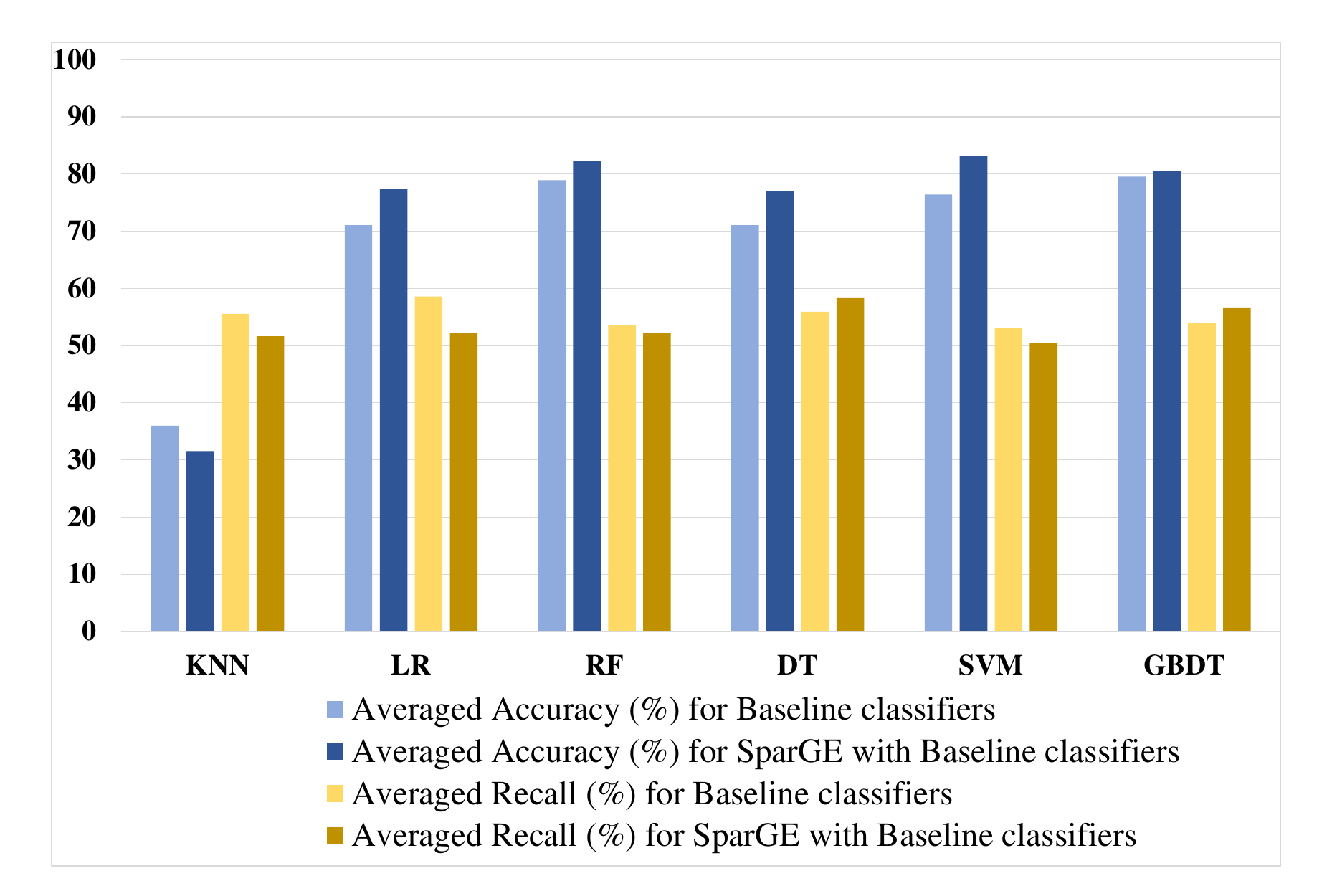}
	\caption{
	 This figure shows the comparison between applying Baseline classifiers on raw \textit{SingHEART} data and on their representations achieved by \textit{SparGE}
		}
\label{fig:dataset1_baseline}
\end{figure}
\begin{figure}[h]
\centering
\includegraphics[scale=0.46]{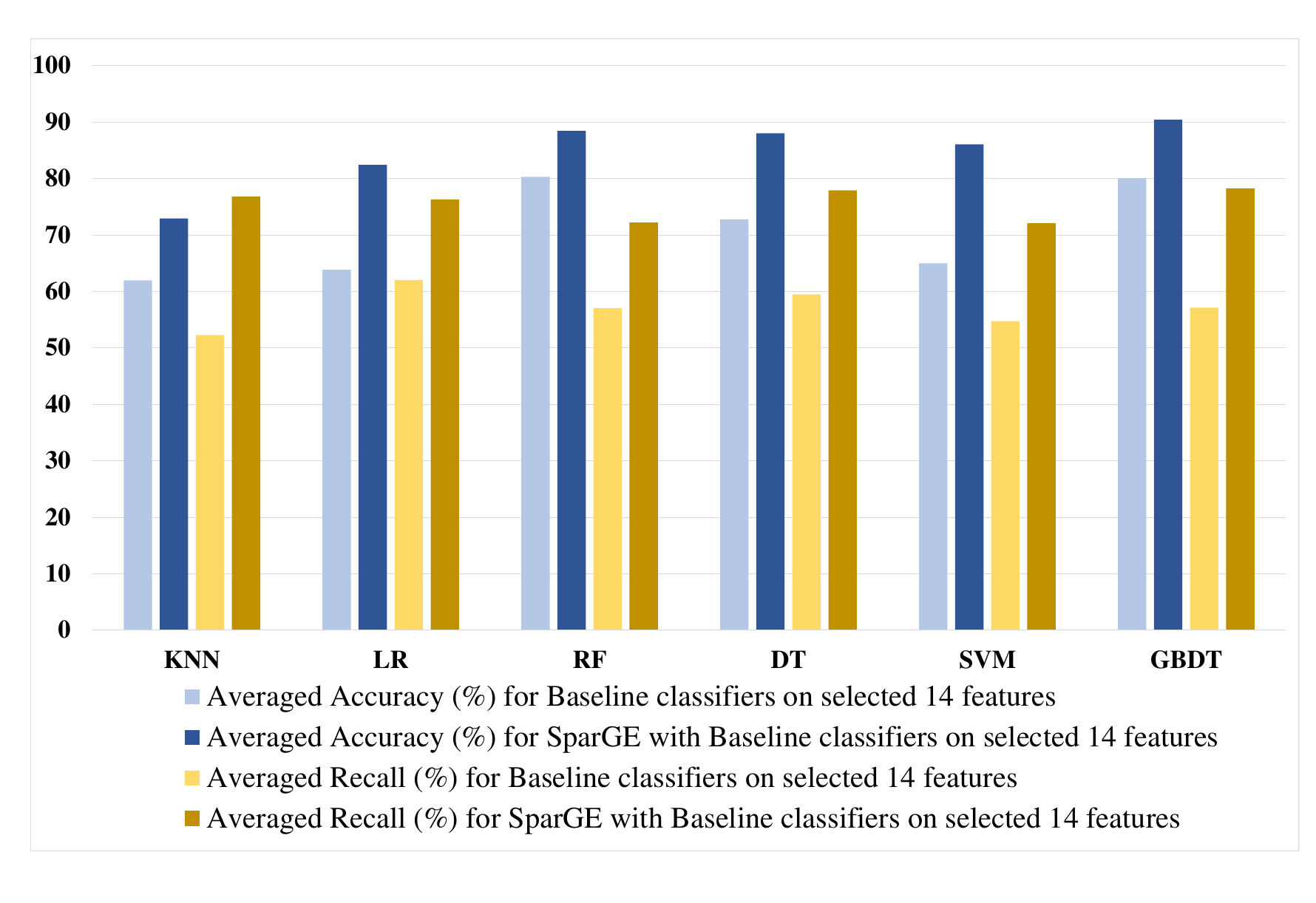}
	\caption{
		Comparing the results on selected $14$ features from \textit{SingHEART} dataset
		}
\label{fig:dataset1_baseline_14features_classifiers}
\end{figure}

This section shows the experimental results on both datasets.
%
By applying the classifiers, shown in Table~\ref{table_classifiers}, on raw \textit{SingHEART} data, we obtain the baseline results.
For comparison, we then apply the classifiers on low-dimensional representations that are achieved by \emph{SparGE}.
Figure~\ref{fig:dataset1_baseline} and Figure~\ref{fig:dataset1_baseline_14features_classifiers} 
compare the their results with same experimental settings.  
The results indicate that our proposed \emph{SparGE} exactly improves the performance on both the accuracy and the recall rate. 
With $14$ selected features from raw dataset, the experiments results are shown in Figure~\ref{fig:dataset1_baseline_14features_classifiers} ,
the \emph{SparGE} still achieves better performance.

\begin{figure}[!htbp]
\centering
\includegraphics[scale=0.46]{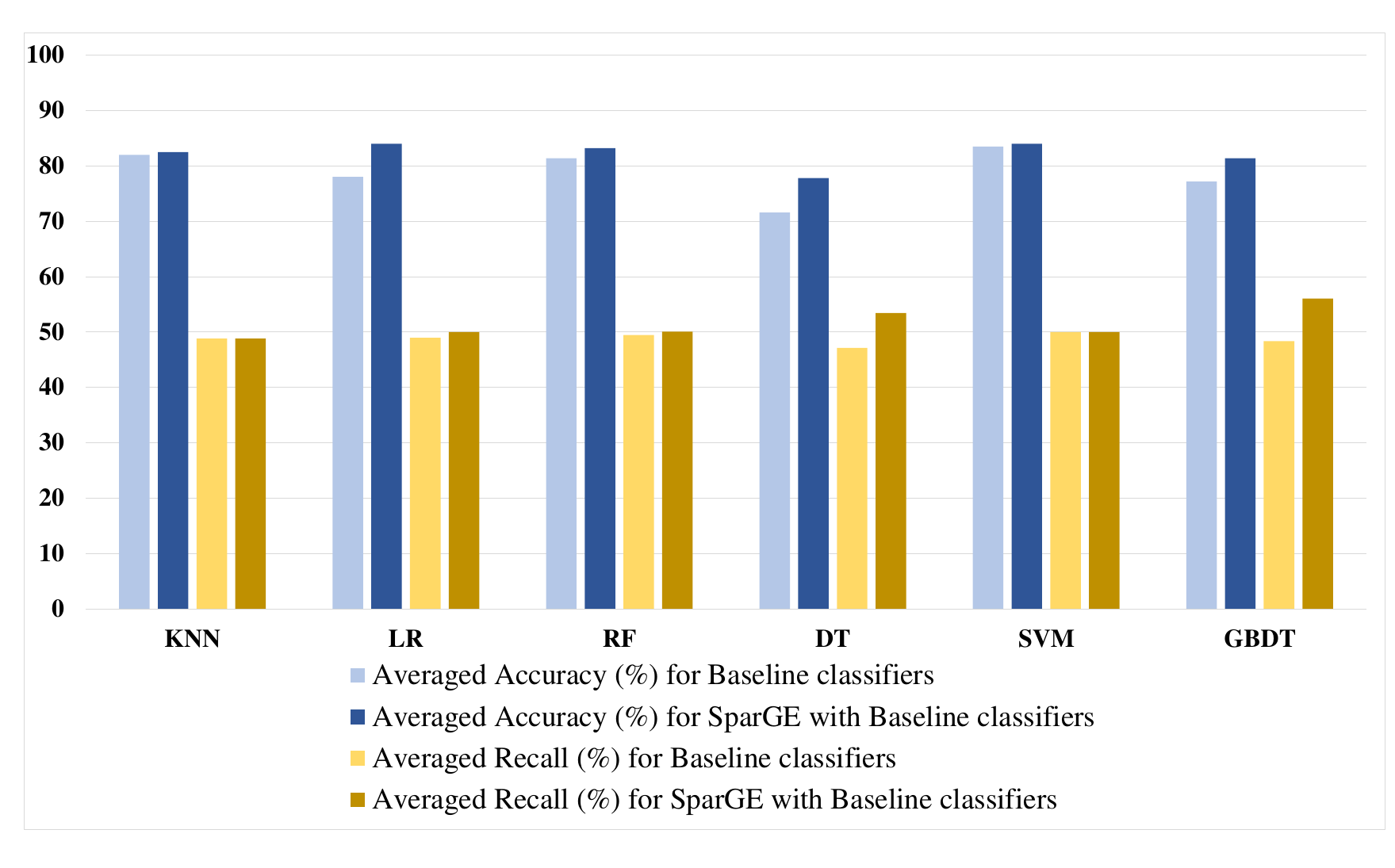}
	\caption{
	Comparing  the  results  on  selected  $17$  fea-tures from MIMIC-III dataset}
\label{fig:dataset2_17features}
\end{figure}

Experimental results on MIMIC-III are collated in Figure~\ref{fig:dataset2_17features} and Table~\ref{table_time}. 
The proposed \emph{SparGE} also show the promising advantages in comparison with baselines. Table~\ref{table_time} shows applying the classifiers on representations achieved by \emph{SparGE} has less computation time.
\begin{table}[h]
\centering
\caption{\label{table_time} 
		Mean time(s) on classifying raw data and the representations achieved by \emph{SparGE}
	}
\begin{tabular}{c|cc}
\hline
 & \textbf{ $\emph{SparGE}$} (s) & \textbf{raw data} (s)\\ \hline
\textbf{KNN} & 0.01627 & 0.056865 \\
\textbf{LR} & 0.018702 & 0.160339 \\
\textbf{RF} & 0.037244 & 0.094995 \\
\textbf{DT} & 0.028631 & 0.257378 \\
\textbf{SVM} & 0.373386 & 1.789594 \\
\textbf{GBDT} & 5.37456 & 11.09086 \\ \hline
\end{tabular}
\end{table}

\section{CONCLUSION}
\label{sec:06}
%

In this work, we presented a joint end-to-end discrimination learning approach, called \emph{SparGE}, 
in order to find and weight similar patients from multi-class EHR data with noise corruption.
At first, sparsity is utilized with low-rank tools to exploit the sparse nature of EHR data with huge amount of missing value.
Then, in order to extract the local similarity from learned sparse coefficients, 
a graph embedding based dimension reduction is applied in sparse domain.
Finally, explicit measurements are constructed in embedded low-dimensional representations. 
By using the differentiability of  sparse regression solutions, 
the \emph{SparGE} could be efficiently solved by a proposed gradient descent algorithm.
%
%
The experimental results show stronger competitive performance of the proposed \emph{SparGE},
 in comparison with other machine learning approaches.
Moreover, the proposed \emph{SparGE} is flexible and can be extended to deeper architectures, applying in more general cases
of dimensionality reduction.

\bibliographystyle{ref.bst}
\bibliography{ref.bib}
%
%





%

%

\newpage

\appendix
\onecolumn
\aistatstitle{SparGE: Sparse Coding-based Patient Similarity Learning via Low-rank Constraints and Graph Embedding: \\
Supplementary Materials}




\section{THE FIRST-ORDER DERIVATIVE OF COST FUNCTION}
 Let $\X = [\x_1, \x_2, \cdots, \x_n]\in \mathbb{R}^{m \times n}$
be a collection of $n$  patients' raw EHR data points, potentially with excessive noises, and let the vector $\x_i = [\x_i^{1}, \x_i^{2}, \cdots, \x_i^{m}]\in \mathbb{R}^{m}$ contain all measurable EHR variables for the $i^{\text{th}}$ patient, with $m$ being the number of all biomarkers.
$\Z := [\z_{1}, \ldots, \z_{n}] \in \mathbb{R}^{m \times n}$ is the low-rank approximated matrix of $\X$. 
We donate by $\D:= [\d_{1}, \ldots, \d_{k}]\in \mathbb{R}^{m\times k}$  a shared dictionary and 
$\Z = \D \Phib$ with $\Phib := [\phib_1, \ldots, \phib_n] \in \mathbb{R}^{k \times n}$.

For each given sample $\z,$ we denote by $\phi^{*}$ the sparse representation with respect to a dictionary $\D^{*}$, we have the unique solution of the sample-wise sparse regression problem as in  Eq. ~\eqref{eq_sparse_regression_solution},

\begin{equation*}
\label{eq_group_sparse_regression}
\begin{split}
f_{\x}(\phib, \Z) := \tfrac{1}{2} \| \z - \D \phib \|_{2}^{2} + g(\phib) + \lambda_1 \|\Z_{\Omega} - \X_{\Omega}\|_F^2 ,
\;\text{with} \; g(\phib) = r_1 \|\phi\|_1 + \frac{r_2}{2} \|\phi\|_F^2,
\end{split}
\end{equation*}

\begin{equation}
\label{eq_sparse_regression_solution}
\phib_{\D}(\x, \z) := \operatorname*{argmin}_{\phib \in \mathbb{R}^{k}} f_{\x}(\phib, \Z).
\end{equation}

where $r1$ and $r2$ are the regularization parameters.

Fix $\mathbf D$, the solution of $\phib^*$ and the derivative of $\phib^*$ to $\D$ are computed by
\begin{equation}
\begin{split}
    \frac{\partial f}{\partial \phib}=0: \phib^{*}&=\mathop{\arg\min}_{\phi \in \mathbb{R}^{k}} \frac{1}{2} \| \z - \D \phib \|_{2}^{2} + r_1 \|\phi\|_1 + \frac{r_2}{2} \|\phi\|_F^2 \\
    &=(\D^{\top}\D + r_2\I)^{-1}(\D^{\top} \z-r_{1} s_{\Lambda})
\end{split}
\end{equation}

\begin{equation}
\label{eq_phitoD}
\begin{split}
    \frac{\partial\phi^*}{\partial \D}
    =& -\D (\D^\top \D + r_2\I)^{-1} (\D^{\top}\z-\lambda_{1}s_{\Lambda}) (\D^\top \D + r_2\I)^{-1} \\
    &- \D (\D^\top \D + r_2\I)^{-1} (\D^{\top}\z-\lambda_{1}s_{\Lambda})^\top (\D^\top \D + r_2\I)^{-1}
    + \z(\D^{\top}\D + r_2\I)^{-1}
\end{split}
\end{equation}

In order to explicitly measure the similarity of patient pairs, 
we already know the \textit{generalized Mahalanobis distance} between $\phi_i$ and $\phi_j$ in sparse domain could be obtained
by computing the Euclidean distance in $\mathbb{R}^{l}$ as follows:
\begin{equation}
\begin{split}
\label{eq_similarity_re}
d^2(\phib_i, \phib_j) 
&=(\phib_i - \phib_j)^\top \P (\phib_i - \phib_j)\\
&=(\mathbf{U}^{\top} \phib_i - \mathbf{U}^{\top} \phib_j)^{\top} (\mathbf{U}^{\top} \phib_i - \mathbf{U}^{\top} \phib_j)\\
& = d^2(\y_i, \y_j)
\end{split}
\end{equation}
where $\mathbf{U} \in \mathbb{R}^{m\times l}$ is an dimension reduction transformation with $l \ll m$, 
$\y_{i} \in \mathbb{R}^{l}$ is a representation in the low-dimensional embedded space. 
Since ${\P}$ is a \textit{Symmetric Positive Semi-Definite}(SPSD) matrix, $\mathbf{U}$ 
could be viewed as a component for approximating ${\P} = \mathbf{U}\mathbf{U}^\top$.
Now, the key is to learn an approximate $\mathbf{U}$, as follows. 

Let $\mathfrak{N}_{k_1}^{+}({\phib}_i)$ denote the set of $k_1$ nearest neighbors which share the same label with ${\phib}_i$, 
and $\mathfrak{N}_{k_2}^{-}({\phib}_i)$ denote the set of $k_2$ nearest neighbors among the data points whose labels 
are different to that of ${\phib}_i$.
We construct two matrices $\Z^{+} := \{z_{ij}^{+}\} \in \mathbb{R}^{n\times n}$
and $\Z^{-} := \{z_{ij}^{-}\} \in \mathbb{R}^{n\times n}$ with
\begin{equation}
	z_{ij}^{+} = \left\{\!\!
	\begin{array}{ll}
		d(\phi_i,\phi_j), & {\phib}_j\in \mathfrak{N}_{k_1}^{+}({\phib}_i) ~\text{or}~ 
		{\phib}_i\in \mathfrak{N}_{k_1}^{+}({\phib}_j), \\
		0, & \text{otherwise},
	\end{array}
	\right.
\end{equation}

\begin{equation}
	z_{ij}^{-} = \left\{\!\!
	\begin{array}{ll}
		d(\phi_i,\phi_j), & {\phib}_j\in \mathfrak{N}_{k_1}^{-}({\phib}_i) ~\text{or}~ 
		{\phib}_i\in \mathfrak{N}_{k_1}^{-}({\phib}_j), \\
		0, & \text{otherwise}.
	\end{array}
	\right.
\end{equation}

\begin{equation}
	y^{+}_{ii} = \sum_{j \neq i} z_{ij}^{+}, \quad\text{and}\quad
	y^{-}_{ii} = \sum_{j \neq i} z_{ij}^{-},
\end{equation}

Then, we have intra-class locality $\mathbf{L}^{+}$ and inter-class locality $\mathbf{L}^{-}$:
\begin{equation}
	\mathbf{L}^{+} = \Y^{+} - \Z^{+}, \quad\text{and}
	\quad \mathbf{L}^{+} = \Y^{-} - \Z^{-}.
\end{equation}

The final jointly cost function for our generic algorithmic framework can be formulated as a problem of the so-called \textit{trace quotient}, i.e.,
\begin{equation}
	\label{eq_trace_quotient_mfa_}
	\operatorname*{\argmin}_{\U, \Z, \D} \;
	\mathcal{L}_{(\mathbf{P}, \mathbf{Z}, \mathbf{D})}.
\end{equation}
with
\begin{equation}
	\label{eq_trace_quotient_mfa}
	\mathcal{L}_{(\mathbf{U}, \mathbf{Z}, \mathbf{D})} := 
	\frac{\operatorname{tr}(\U^{\top} {\Phib_{\X}(\Z, \D)}  \mathbf{L}^{+} {\Phib_{\X}(\Z, \D)}^\top \U)}{
		\operatorname{tr}(\U^{\top}{\Phib_{\X}(\Z, \D)} 
		\mathbf{L}^{-} {\Phib_{\X}(\Z, \D)}^\top \U) },
\end{equation}

The derivatives of $\mathcal{L}_{(\mathbf{U}, \mathbf{Z}, \mathbf{D})}$ with respect to $\P$ and $\D$ are computed as
\begin{equation}
\begin{split}
    \label{eq_LtoU}
	\frac{\partial \mathcal{L}}{\partial \U} =
	2[ \frac{\Phib \mathbf{L}^+ \Phib^T \U + \Phib^\top \U \U^\top \Phib \frac{\partial \mathbf{L}^+}{\partial \U} }{\operatorname{tr}(\U^{\top} \Phib_{\X} \mathbf{L}^- \Phib^\top \U)} - 
	&\frac{(\Phib \mathbf{L}^- \Phib^T \U + \Phib^\top \U\U^\top \Phib \frac{\partial \mathbf{L}^-}{\partial \U}) \operatorname{tr}(\U^{\top} \Phib \mathbf{L}^{+} \Phib^\top \U)}
	{\operatorname{tr}^2(\U^{\top} \Phib_{\X} \mathbf{L}^- \Phib^\top \U)} ],
\end{split}
\end{equation}

\begin{equation}
\begin{split}
    \label{eq_LtoD}
	\frac{\partial \mathcal{L}}{\partial \D} = 	\frac{\partial \mathcal{L}}{\partial \Phib}\frac{\partial \Phib}{\partial \D}=
	2[ \frac{\U\U^{\top}\Phib (\mathbf{L}^+)^\top}{\operatorname{tr}(\U^{\top} \Phib_{\X} \mathbf{L}^- \Phib^\top \U)} - 
    \frac{\U\U^{\top}\Phib (\mathbf{L}^-)^\top \operatorname{tr}(\U^{\top} \Phib_{\X} \mathbf{L}^+ \Phib^\top \U)}{\operatorname{tr}^2(\U^{\top} \Phib_{\X} \mathbf{L}^- \Phib^\top \U)} ] \frac{\partial \Phib}{\partial \D}.
\end{split}
\end{equation}
Therein, $\frac{\partial \mathbf{L}^+}{\partial \U}$ is computed by
\begin{equation}
	\frac{\partial \mathbf{L}^+}{\partial \U} =\sum^{n}_{i,j=1} \left\{\!\!
	\begin{array}{ll}
		\frac{\partial L_{ij}^+}{\partial \U}, & {L_{ij}^+}\neq0
		, \\
		0, & \text{otherwise},
	\end{array}
	\right.
\end{equation}
\begin{equation}
	\frac{\partial L_{ij}^+}{\partial \U} =\left\{\!\!
	\begin{array}{ll}
		-\frac{\partial \z_{ij}^+}{\partial \U}, & i\neq j, \\
		\sum_{t=1}^{n}\frac{\partial \z_{it}^+}{\partial \U}. & i=j
	\end{array}
	\right.
\end{equation}

\begin{equation}
\frac{\partial \z_{ij}^+}{\partial \U} = 2(\phib_i - \phib_j)^\top(\phib_i - \phib_j)  \U
\end{equation}
with
\begin{equation}
\begin{split}
\frac{\partial \Phib}{\partial \D} = \sum^{n}_{i=1} \frac{\partial \phib_i}{\partial \D}
\end{split}
\end{equation}
where $\frac{\partial \phib_i}{\partial \D}$ is computed by Eq.~\eqref{eq_phitoD}.
The similar computation is for $\frac{\partial \mathbf{L}^-}{\partial \U}$.

With $\frac{\partial \mathcal{L}}{\partial \D}$, $\frac{\partial \mathcal{L}}{\partial \U}$ at hand, the updates for 
$\D$ and $\U$ are described in Algorithm~$1$,
where the projections from ${\hat{\D}}^{(j+1)}, {\hat{\U}}^{(j+1)}$ onto ${{\D}}^{(j+1)} \in \mathcal{D}, {{\U}}^{(j+1)} \in \mathcal{U}$
are computed by the unit normalization and the QR decomposition, respectively.

\section{EXPERIMENTS}
In the experiments, we set $70\%$ as training data, and $30\%$ as testing data. 
The following charts are supplementary to the two datasets used in the experiments.
%
\subsection{SingHeart Dataset}
Table~\ref{table_singhealth} shows biomarkers in \emph{SingHEART} dataset and the number of corresponding variables.
\begin{table}[htbp]
\centering
	\caption{\label{table_singhealth} 
	\emph{SingHEART} biomarkers from \textit{Singapore Health Services, SingHealth}, with multi-source sensors \cite{yap2019harnessing_BCD}
	}
\begin{tabular}{cc}
\hline
\textbf{Measurement method} & \textbf{Number of variables} \\ \hline
Calcium Score & 7 \\ \hline
APB Biofourmis & 101 \\ \hline
Lab Tests & 35 \\ \hline
MRI Features & 29 \\ \hline
Proforma & 88 \\ \hline
General Questionnaire & 189 \\ \hline
Lifestyle Questionnaire & 18 \\ \hline
\end{tabular}
\end{table}


\subsection{MIMIC-III Dataset}
Considering the large size of MIMIC-III dataset, after selecting samples, 
17 features are selected from them. The first column of Table ~\ref{table_mimic3} 
shows the $17$ features corresponding to clinical variables. In Table ~\ref{table_mimic3}, 
the source tables of a variable from MIMIC-III database are shown in the second column. 
There are two main source tables, 'chartevents' and 'Labevents'.
The third column lists the “normal” values we used in our baselines during the imputation step, 
and the fourth column describes how baselines and our models treat the variables.

\begin{table}[h]
\centering
\caption{\label{table_mimic3}The $17$ selected clinical variables. The second column shows the source table(s) of a variable from MIMIC-III database. 
	The third column lists the “normal” values we used in our baselines during the imputation step, and the fourth column describes 
	how models treat the variables.}
\begin{tabular}{cccc}
\hline
\textbf{Variables} & \textbf{MIMIC-III table} & \textbf{Impute value} & \textbf{Modeled as} \\ \hline
Capillary refill rate & chartevents & 0.0 & categorical \\
Diastolic blood pressure & chartevents & 59.0 & continuous \\
Fraction inspired oxygen & chartevents & 0.21 & continuous \\
Glasgow coma scale eye opening & chartevents & 4 spontaneously & categorical \\
Glasgow coma scale motor response & chartevents & 6 obeys commands & categorical \\
Glasgow coma scale total & chartevents & 15 & categorical \\
Glasgow coma scale verbal response & chartevents & 5 oriented & categorical \\
Glucose & chartevents, Labevents & 128.0 & continuous \\
Heart Rate & chartevents & 86 & continuous \\
Height Rate & chartevents & 170.0 & continuous \\
Mean blood pressure & chartevents & 77.0 & continuous \\
Oxygen saturation & chartevents, Labevents & 98.0 & continuous \\
Respiratory rate & chartevents & 19 & continuous \\
Systolic blood pressure & chartevents & 118.0 & continuous \\
Temperature & chartevents & 36.6 & continuous \\
Weight & chartevents & 81.0 & continuous \\
PH & chartevents, Labevents & 7.4 & continuous \\ \hline
\end{tabular}
\end{table}

The value of $17$ channels are given including e.g., normal values which provides a normal standard, and $5$ categorical channels, which are capillary refill rate, glasgow coma scale eye opening, glasgow coma scale motor response, glasgow coma scale total and glasgow coma scale verbal response. 

For glasgow coma scale eye opening, we divide it into 4 grades according to medical standards. 
Respectively, 4 points means patients can open their eyes spontaneously. 3 points means the patient can open his eyes when called. 
2 points means the patient can open his eyes when pained. 1 point means the patient has no response. 

We divide glasgow coma scale motor response into 6 grades according to medical standards. 
Respectively, 6 points means that the patient can obey commands to make a motor response, which demonstrates the patient is a normal person. 
5 points indicate that stimulation of the patient can localize pain. 
4 points means the patient has flex withdraws. 3 points means the patient has abnormal flexion. 
2 points means the patient has abnormal extension. 1 point means the patient has no response. 

As for glasgow coma scale verbal response, we divide it into 6 grades according to medical standards.  
Respectively, 5 points indicates that the patient is organized. 4 points means the patient is confused. 
3 points means that the patient speaks inappropriate words. 2 points means the patient can only make incomplete sounds. 
1 point means the patient has no response.

\vfill

\end{document}